# MAMI: Multi-Attentional Mutual-Information for Long Sequence Neuron Captioning

Alfirsa Damasyifa Fauzulhaq*

Faculty of Computer Science, Brawijaya University, Indonesia, alfirsaafauzulh@gmail.com

Wahyu Parwitayasa

Faculty of Computer Science, Brawijaya University, Indonesia, parwita@student.ub.ac.id

Joseph Ananda Sugihdharma

Faculty of Computer Science, Brawijaya University, Indonesia, josephananda88@student.ub.ac.id

M. Fadli Ridhani

Faculty of Computer Science, Brawijaya University, Indonesia, mfadliridhani @student.ub.ac.id

Novanto Yudistira

Faculty of Computer Science, Brawijaya University, Indonesia, yudistira@ub.ac.id

Neuron labeling is an approach to visualize the behaviour and respond of a certain neuron to a certain pattern that activates the neuron. Neuron labeling extract information about the features captured by certain neurons in a deep neural network, one of which uses the encoder-decoder image captioning approach. The encoder used can be a pretrained CNN-based model and the decoder is an RNN-based model for text generation. Previous work, namely MILAN (**M**utual **I**nformation-guided **L**inguistic **A**nnotation of **N**euron), has tried to visualize the neuron behaviour using modified Show, Attend, and Tell (SAT) model in the encoder, and LSTM added with Bahdanau attention in the decoder. MILAN can show great result on short sequence neuron captioning, but it does not show great result on long sequence neuron captioning, so in this work, we would like to improve the performance of MILAN even more by utilizing different kind of attention mechanism and additionally adding several attention result into one, in order to combine all the advantages from several attention mechanism. Using our compound dataset, we obtained higher BLEU and F1-Score on our proposed model, achieving 17.742 and 0.4811 respectively. At some point where the model converges at the peak, our model obtained BLEU of 21.2262 and BERTScore F1-Score of 0.4870.

CCS COibNCEPTS • Insert your first CCS term here • Insert your second CCS term here • Insert your third CCS term here

**Additional Keywords and Phrases:** Image Captioning, Deep Neural Network, Computer Vision, Text Generation

**ACM Reference Format:**
First Author's Name, Initials, and Last Name, Second Author's Name, Initials, and Last Name, and Third Author's Name, Initials, and Last Name. 2018. The Title of the Paper: ACM Conference Proceedings Manuscript Submission Template: This is the subtitle of the paper, this document both explains and embodies the submission format for authors using Word. In Woodstock '18: ACM Symposium

---

* Place the footnote text for the author (if applicable) here.

## 1 INTRODUCTION

It is possible to understand the complex behavior of a network by studying the individual properties of each neuron that makes up the network. Previous studies on automatic visualization and categorization of neurons have found that each neuron in a neural network has a sensitivity to a certain pattern in the input data. It includes identifying the features and functions implemented by each neuron in the network [1]. After going through several layers of the neural network, several units will be activated for certain sections and categories of objects [2]. Single neurons in neural networks have also been found to have the ability to encode sentiments or emotions in language data [3] and the biological role in computational chemistry [4].

There are limited behavior characterizations of individual neurons. Visualization based approach [5]–[8] give annotation work to human users, so it can't be used for large-scale analysis. Existing auto-labeling techniques [2], [9] the candidate neuron labels in previous studies were predefined by the researchers on a small subset of neurons, not allowing novel or unexpected behaviors to emerge.

Research from [10] give a solution by labeling neurons with expressive, compositional, and open annotations in the form of natural language descriptions. And called it **M**utual **I**nformation-guided **L**inguistic **A**nnotation of **N**euron (MILAN).

MILAN uses Long Short Term Memory (LSTM) and Bahdanau Attention for its decoder [10]. LSTMs require many parameters and significant computational resources, larger LSTM models will have more parameters to train, in addition to their susceptibility to vanishing gradient and exploding gradient problems [11]. However, MILAN still didn't perform well in captioning datasets with the longest maximum number of words. In datasets with the longest maximum number of words, all captions in the dataset will be padded according to the maximum word count. For example, if the maximum word count in the dataset is 47 words, all caption data in the dataset will be padded to a length of 47 words. The objective of this research is to enhance the capability of MILAN in handling datasets with longer maximum word counts by employing a Multi Attention Mechanism.

## 2 MULTI-ATTENTIONAL NEURON LABELING

### 2.1 Dataset

On this study, the Places365 dataset from Milannotations were used. There are 3 variants of Places365 which used and trained for our proposed model, namely dataset from AlexNet, ResNet-152, and BigGAN. The ImageNet dataset which used in Milannotations were unavailable to be accessed and downloading all ImageNet is time consuming and require a lot of memory space. The characteristics of the dataset is shown in Table 1.

Table 1: Dataset Characteristics

| Data Characteristics | Alexnet | ResNet-152 | BigGAN |
|---|---|---|---|
| Data Amount | 1376 | 3904 | 4992 |
| Longest Sequence Length | 21 | 47 | 39 |
| Average Sequence Length | 4 | 5 | 4 |
| Most Appeared Seq. Length | 5 | 4 | 4 |
| Occurences of Most Appeared Length | 896 times | 2577 times | 2838 times |
| Top 10 Word Occurences | [objects, and, the, of, colored, white, or, with, lines, black] | [a, and, the, of, in, on, people, objects, space, with] | [and, of, items, merchandise, store, a, the, areas, people, in] |

## 2.2 Mutual-Information-Guided Linguistic Annotation of Neurons (MILAN)

In the formula, MILAN($f_i$) represents the natural language description of the neuron ($f_i$) generated by the MILAN algorithm. The formula is used to find the description (d) that maximizes the point-wise mutual information (pmi) between the description and the exemplar set ($E_i$) of the neuron ($f_i$). The (pmi) is calculated as the difference between the log probability of the description given the exemplar set and the log probability of the description alone. The MILAN formula is as follows:

$$MILAN(f_i) = argmax\ pmi(d; E_i) = argmax\ log\ p(d|E_i) - log\ p(d) \qquad (1)$$

Where $pmi(d; E_i)$ is the conditional probability of the description given the exemplar set ($E_i$), and $p(d)$ is the marginal probability of the description. The MILAN algorithm uses this formula to search for a natural language description that is highly specific to the pattern of activation of the neuron ($f_i$) on input images [10].

## 2.3 Bahdanau Attention

Attention is widely used in modern neural architectures [12], In particular, it is the heart of the Transformer architecture [13], In the context of neural machine translation (NMT), the main purpose of Bahdanau Attention is to assist the model in learning the dependency between the words in the source sentence and the words in the target sentence. By using attention, the model can adjust the weight or importance of each word in the source sentence when generating each word in the target sentence. The bahdanau attention formula is as follows:

$$Attention: \alpha_{ts} = \frac{exp(e_{ts})}{\sum_{s'} exp(e_{ts'})}, with: e_{ts} = v_a^T tanh(W_a[h_t; h_s]) \qquad (2)$$

Where $h_t$ is the attention weight between the target representation vectors, $h_s$ is the source representation vector which is calculated using a linear combination of the two vectors with the learning matrix $W_a$, followed by the tanh function and dot product operation with the learning vector $v_a$ [14].

## 2.4 Luong Attention

There are mainly two types of attention, namely global attention and local attention. What differs both between them is the placement of the attention mechanism, whether it is placed only at a few parts of the sequence or at the whole sequence. In Bahdanau attention mechanism [15], the forward and backward source hidden states in the bi-directional encoder and target hidden states in the non-stacked unidirectional encoder were combined. Meanwhile, in the Luong attention mechanism, the global attention mechanism is simplified by using hidden states on the top of the LSTM layer in the encoder and decoder. The luong attention formula is as follows:

$$Attention: \alpha_{ts} = \frac{exp(e_{ts})}{\sum_{s'} exp(e_{ts'})}, with: e_{ts} = h_t^T W_a h_s \qquad (3)$$

Where $h_t$ is the attention weight between the target representation vectors, $h_s$ is the source representation vector which is calculated using the dot product operation between the two vectors [16].

### 2.5 Self Attention

Self-attention, also known as attention mechanism in Natural Language Processing (NLP) neural networks such as Transformer, is a critical component that allows the model to focus on relevant parts of the input text. In the context of NLP, self-attention allows the model to learn the context representation of words in the text. The self-attention formula is as follows:

$$Attention: (Q, K, V) = softmax\left(\frac{QK^T}{\sqrt{d_k}} V\right) \qquad (4)$$

Where $Q$ is the query matrix, $K$ is the key matrix, $V$ is the value matrix, and $d_k$ is the dimension of the matrix [13].

### 2.6 Multi Attention

In this study, we proposed MAMI (Multi-Attentional MILAN for Neuron Labeling). MAMI is an attention mechanism that consisted of multiple attentions, which are Bahdanau Attention, Luong Attention, and Self Attention. The results from that multiple attentions are summed up and then used by decoder as an additional information in order to generate caption. The illustration of our proposed model is shown in Figure 1 and the Multi Attention formula is as follows :

$$Attention: v_a^T tanh(W_a[h_t; h_s]) + h_t^T W_a h_s + softmax\left(\frac{QK^T}{\sqrt{d_k}} V\right) \qquad (5)$$

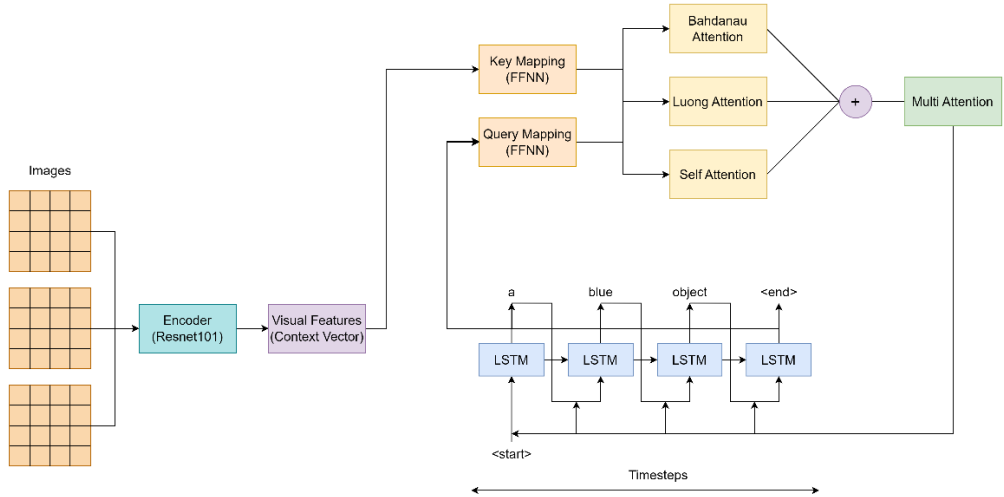

Figure 1: MAMI Illustration

## 2.7 Experiments

On the conducted experiments in this study, we changed some hyperparameters of the model, such as batch size and learning rate due to the limitations on our resource. Besides, since the training and testing data split from the original paper is not disclosed, we create our own data split and retrain all the model including the baseline to maintain same configurations and environment on all trained model. We tested different kind of attention mechanism on the model, in order to find out the increasing BLEU and BERT Score from the generated caption as an output of the model.

## 2.8 Evaluation

Evaluation metrics used in this model namely, BLEU score [17] and BERTScore [18]. BLEU score used to measure the quality of the generated caption from the model based on n-grams approaches. Formula to calculate BLEU score is shown in Equation 6. Meanwhile BERTScore measure the similarity between words token in the references $x_i$ and candidates $\hat{x}_j$ using pre-trained BERT contextual embedding. BERTScore uses point-wise cosine similarity to measure the similarity between references and candidates value as in equation 7. From BERTScore, we calculate the precision, recall, and f1-score from each evaluation metrics using equation 8-12. We specifically used the f1-score value from the BERTScore since it is the value shown in the previous research and this metric can be used to compare their models with ours.

$$BLEU = bp * e^{\frac{\sum_{\text{n-grams} \in \hat{y}} Count_{\text{clip}}(\text{n-gram})}{\sum_{\text{n-grams} \in \hat{y}} Count(\text{n-gram})}} \quad (6)$$

$$similarity(x_i, \hat{x}_j) = \frac{x_i^T \hat{x}_j}{|x_i||\hat{x}_j|} \quad (7)$$

$$RBERT = \frac{1}{|x|} \sum_{x_i \in x} max_{\hat{x}_j \in \hat{x}} x_i^T \hat{x}_j \quad (8)$$

$$PBERT = \frac{1}{|\hat{x}|} \sum_{\hat{x}_j \in x} max_{x_i \in x} x_i^T \hat{x}_j \quad (11)$$

$$FBERT = 2 \frac{P_{BERT} \cdot R_{BERT}}{P_{BERT} + R_{BERT}} \quad (12)$$

## 3 RESULTS AND DISCUSSIONS

Our results is shown in Table 2. From these result, we could see the F1 BERTScore from our proposed model, namely LSTM-Multi Attention achieved higher result compared with the baseline model and re-trained baseline model. Based on the result, not all the result from our proposed LSTM-Multi Attention surpassed all other models in every dataset we used. For example, in AlexNet dataset, we could see the baseline model produce the highest F1-Score, LSTM-Self Attention performs best in ResNet-152 dataset, meanwhile LSTM-Luong Attention performs best in BigGAN dataset. By comparing the BLEU score, we could see LSTM-Multi Attention performance is dominating almost on every dataset, leaving only small difference rather than LSTM-Luong Attention in AlexNet dataset and LSTM-Self Attention in BigGAN dataset.

Based on the results we achieved, we did some analysis of our model's performances and looked out for an answer to "Why is our LSTM-Multi Attention model not surpassing all other models in all datasets?". From our assumptions, if the datasets sorted from the most complex dataset to the most simple dataset (ResNet-152 - BigGAN - AlexNet), also the longest sequence and most appeared word were brought into consideration, we concluded LSTM-Multi Attention performs really well in the most complex dataset, namely ResNet-152 and achieve BLEU with very small differences from the best model in the simplest or easiest dataset, namely AlexNet.

It also can be concluded that we don't need a complex model to learn the AlexNet dataset really well. To prove our hypothesis, we train those proposed models to the compound dataset of Places365 (ResNet-152 + AlexNet + BigGAN) with a possibility of our LSTM-Multi Attention model learning the complex and big dataset really well and achieving the best performance in this dataset. From the experiments conducted, we could see that our proposed model, namely LSTM-Multi Attention could outperform other models' performance, such as LSTM-Bahdanau, LSTM-Luong, and LSTM-Self Attention models in the BLEUScore and BERTScore evaluation metrics. Additionally, in every training process, we saved those proposed models at the highest validation BLEU Score (Compound-Best-BLEU) and tested those models to know the value of the BLEUScore and BERTScore as shown in Figure 2 below. We concluded that the LSTM-Multi Attention model still shows its performance and outperforms other models' performance.

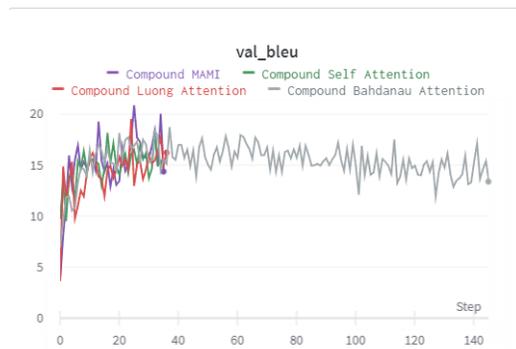

Figure 2: Model BLEU Results

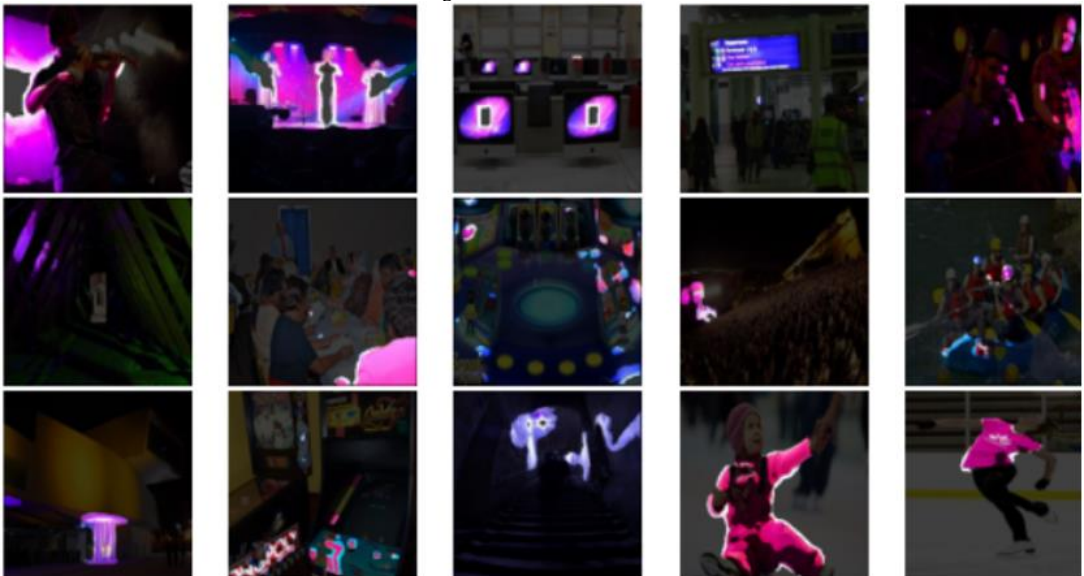

Annotation: image with colorful image in support of the kind of images that have different pictures underneath and the picture tell different appealing stories there is shapes at the center of each images
MILAN (baseline): Pink colored objects
**MAMI**: **Fluorescent pink and purple objects**

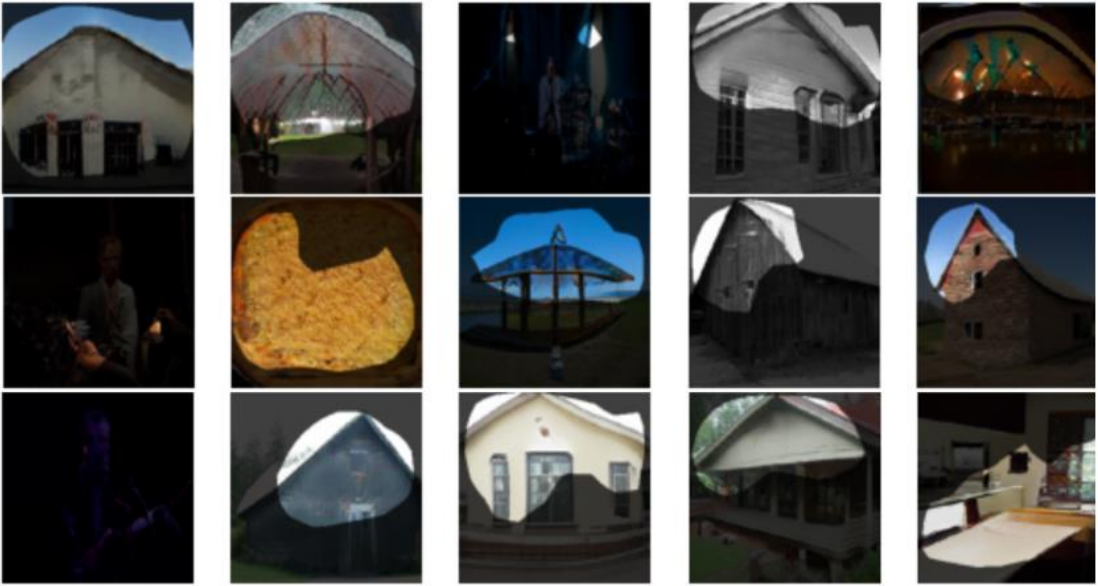

Annotation         : the tops of buildings, windows, part of the middle of a pizza, tops of chairs and top of a table, the arm of a shirt, and some sort of decoration on a wall
MILAN (baseline) : Roofs and roofs
**MAMI              : Pizza and roofs**

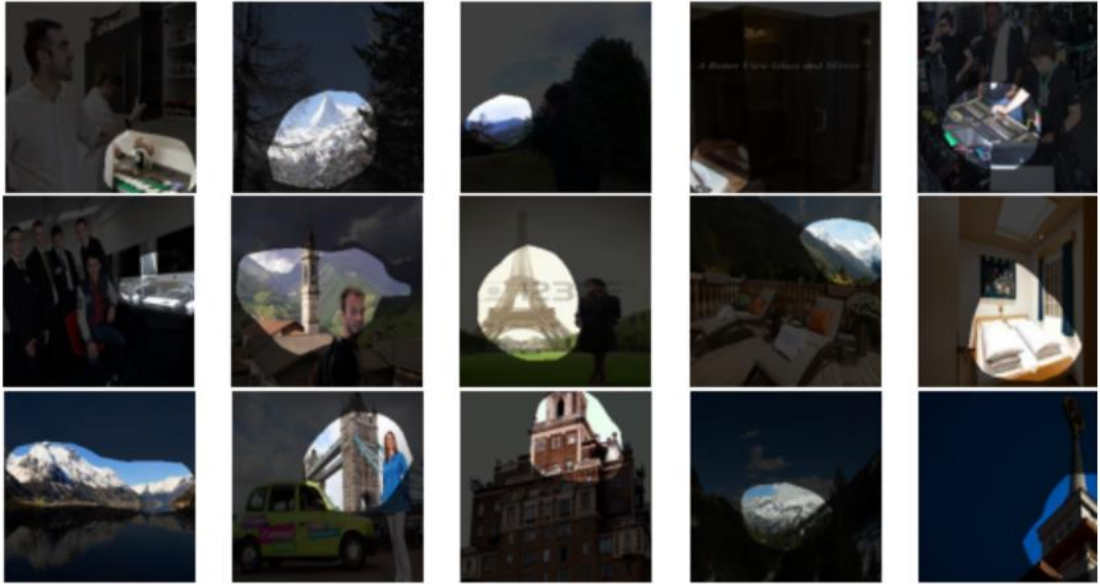

Annotation         : shown are mountains, a do booth, a double bed, a woman in a blue shirt in front of a bridge, a brick building, the eiffel tower and a mans face in front of of a tall slim building
MILAN (baseline) : Towers
**MAMI              : Towers and mountains**

Figure 3: MAMI Caption Output

Table 2: BLEU and BERTScore on Each Dataset

| Dataset | Model | BLEU | BERTScore | | |
| --- | --- | --- | --- | --- | --- |
| | | | Precision | Recall | F1-Score |
| AlexNet | Bahdanau (Baseline) | - | - | - | **0.4700** |
| | Bahdanau retrained | 12.3666 | 0.4740 | 0.3589 | 0.4026 |
| | Luong Attention | **12.8666** | 0.4996 | 0.3952 | 0.4373 |
| | Self Attention | 11.1016 | 0.4748 | 0.3642 | 0.4100 |
| | MAMI | 12.8611 | 0.4659 | 0.3892 | 0.4145 |
| ResNet-152 | Bahdanau (Baseline) | - | - | - | 0.2800 |
| | Bahdanau retrained | 3.8679 | 0.3313 | 0.3178 | 0.3107 |
| | Luong Attention | 5.6136 | 0.4337 | 0.3008 | 0.3521 |
| | Self Attention | 6.7876 | 0.4603 | 0.2947 | **0.3607** |
| | MAMI | **8.5397** | 0.4430 | 0.2910 | 0.3498 |
| BigGAN | Bahdanau (Baseline) | - | - | - | 0.5200 |
| | Bahdanau retrained | 30.2698 | 0.5894 | 0.4940 | 0.5308 |
| | Luong Attention | 24.5812 | 0.5834 | 0.5241 | **0.5458** |
| | Self Attention | **34.8311** | 0.5839 | 0.5025 | 0.5334 |
| | MAMI | 32.2052 | 0.5832 | 0.5023 | 0.5328 |
| Compound | Bahdanau (Baseline) | - | - | - | - |
| | Bahdanau retrained | 14.5538 | 0.5341 | 0.4340 | 0.4733 |
| | Luong Attention | 16.0248 | 0.5293 | 0.4507 | 0.4784 |
| | Self Attention | 15.7811 | 0.5204 | 0.4425 | 0.4712 |
| | MAMI | **17.7420** | 0.5455 | 0.4394 | **0.4811** |
| Compound-Best-BLEU | Bahdanau (Baseline) | - | - | - | - |
| | Bahdanau retrained | 20.0595 | - | - | 0.4685 |
| | Luong Attention | 20.2451 | - | - | 0.4747 |
| | Self Attention | 18.6446 | - | - | 0.4764 |
| | MAMI | **21.2262** | - | - | **0.4870** |

## 4 CONCLUSIONS

From the experiments that we conducted, we conclude that our proposed model, using LSTM and Multi-Attention or also known as MAMI (Multi-Attentional MILAN for Neuron Labeling), we could obtained BLEU Score of 17.742 using compound dataset and BERTScore 0.4811, outperforming the baseline of MILAN which use LSTM and Bahdanau attention. On the peak of our model convergency, our model could reach BLEU Score of 21.2262 and BERTScore of 0.4870.

**ACKNOWLEDGMENTS**

**REFERENCES**

[1] D. Erhan, Y. Bengio, A. Courville, and P. Vincent, "Visualizing higher-layer features of a deep network," *Bernoulli*, no. 1341, pp. 1–13, 2009, [Online]. Available: http://igva2012.wikispaces.asu.edu/file/view/Erhan+2009+Visualizing+higher+layer+features+of+a+deep+network.pdf

[2] D. Bau, B. Zhou, A. Khosla, A. Oliva, and A. Torralba, "Network Dissection: Quantifying Interpretability of Deep Visual Representations," Apr. 2017, [Online]. Available: http://arxiv.org/abs/1704.05796

[3] A. Radford, R. Jozefowicz, and I. Sutskever, "Learning to Generate Reviews and Discovering Sentiment," Apr. 2017, [Online]. Available: http://arxiv.org/abs/1704.01444

[4] K. Preuer, G. Klambauer, F. Rippmann, S. Hochreiter, and T. Unterthiner, "Interpretable Deep Learning in Drug Discovery," Mar. 2019, [Online]. Available: http://arxiv.org/abs/1903.02788

[5] M. D. Zeiler and R. Fergus, "Visualizing and Understanding Convolutional Networks," Nov. 2013, [Online]. Available: http://arxiv.org/abs/1311.2901

[6] R. Girshick, J. Donahue, T. Darrell, and J. Malik, "Rich feature hierarchies for accurate object detection and semantic segmentation," Nov. 2013, [Online]. Available: http://arxiv.org/abs/1311.2524

[7] A. Karpathy, J. Johnson, and L. Fei-Fei, "Visualizing and Understanding Recurrent Networks," Jun. 2015, [Online]. Available: http://arxiv.org/abs/1506.02078

[8] A. Mahendran and A. Vedaldi, "Understanding Deep Image Representations by Inverting Them," Nov. 2014, [Online]. Available: http://arxiv.org/abs/1412.0035

[9] J. Mu and J. Andreas, "Compositional Explanations of Neurons," Jun. 2020, [Online]. Available: http://arxiv.org/abs/2006.14032

[10] E. Hernandez, S. Schwettmann, D. Bau, T. Bagashvili, A. Torralba, and J. Andreas, "Natural Language Descriptions of Deep Visual Features," no. 2, pp. 1–21, 2022, doi: https://doi.org/10.48550.

[11] S. Bai, J. Z. Kolter, and V. Koltun, "An Empirical Evaluation of Generic Convolutional and Recurrent Networks for Sequence Modeling," Mar. 2018, [Online]. Available: http://arxiv.org/abs/1803.01271

[12] M. N. Rabe and C. Staats, "Self-attention Does Not Need $O(n^2)$ Memory," Dec. 2021, [Online]. Available: http://arxiv.org/abs/2112.05682

[13] A. Vaswani *et al.*, "Attention is all you need," *Adv. Neural Inf. Process. Syst.*, vol. 2017-Decem, no. Nips, pp. 5999–6009, 2017.

[14] R. Bal and S. Sinha, "Modelling Bahdanau Attention using Election methods aided by Q-Learning," Nov. 2019, [Online]. Available: http://arxiv.org/abs/1911.03853

[15] D. Bahdanau, K. H. Cho, and Y. Bengio, "Neural machine translation by jointly learning to align and translate," *3rd Int. Conf. Learn. Represent. ICLR 2015 - Conf. Track Proc.*, pp. 1–15, 2015.

[16] M.-T. Luong, H. Pham, and C. D. Manning, "Effective Approaches to Attention-based Neural Machine Translation," Aug. 2015, [Online]. Available: http://arxiv.org/abs/1508.04025

[17] K. Papineni, S. Roukos, T. Ward, and W.-J. Zhu, "BLEU: a method for automatic evaluation of machine translation," *Acl*, no. February, pp. 311–318, 2001, doi: 10.3115/1073083.1073135.

[18] T. Zhang, V. Kishore, F. Wu, K. Q. Weinberger, and Y. Artzi, "BERTScore: Evaluating Text Generation with BERT," pp. 1–43, 2019, [Online]. Available: http://arxiv.org/abs/1904.09675